\documentclass{article}

\usepackage[nonatbib,preprint]{neurips_2025}

\usepackage[utf8]{inputenc} 
\usepackage[T1]{fontenc}    
\usepackage{hyperref}       
\usepackage{url}            
\usepackage{booktabs}       
\usepackage{amsfonts}       
\usepackage{nicefrac}       
\usepackage{microtype}      

\usepackage[ruled,vlined]{algorithm2e}
\usepackage{makecell}
\usepackage{graphicx}
\usepackage{amsmath, bm}
\usepackage{tabularx,booktabs}
\usepackage{adjustbox}
\usepackage{multirow}
\usepackage[dvipsnames,table,xcdraw]{xcolor}
\definecolor{LightCyan}{rgb}{0.88,1,1}
\usepackage{array}

\usepackage{subcaption}
\usepackage{indentfirst}
\usepackage{pifont}
\usepackage{hyperref}
\hypersetup{
    colorlinks=true,
    citecolor=blue,
    }
\usepackage{amssymb}
\usepackage{tikz}
\usepackage{xcolor}

\newcolumntype{C}[1]{>{\centering\arraybackslash}p{#1}}
\newcolumntype{L}[1]{>{\arraybackslash}p{#1}}

\newcommand{\etal}{\textit{et al}}
\newcommand{\ie}{\textit{i.e.}}
\newcommand{\eg}{\textit{e.g.}}
\newcommand{\etc}{\textit{etc}}

\title{BenchDepth: Are We on the Right Way to Evaluate Depth Foundation Models?}

\author{
Zhenyu Li$^{1,}$$^2$, Haotong Lin$^{2,}$$^3$, Jiashi Feng$^2$, Peter Wonka$^1$, Bingyi Kang$^2$ \\
$^1$KAUST, $^2$ByteDance Seed, $^3$Zhejiang University \\
\small\url{https://zhyever.github.io/benchdepth/} \\
}

\begin{document}

\maketitle


\begin{abstract}

Depth estimation is a fundamental task in computer vision with diverse applications. Recent advancements in deep learning have led to powerful depth foundation models (DFMs), yet their evaluation remains challenging due to inconsistencies in existing protocols. Traditional benchmarks rely on alignment-based metrics that introduce biases, favor certain depth representations, and complicate fair comparisons. 
In this work, we propose \textbf{\textit{BenchDepth}}, a new benchmark that evaluates DFMs through five carefully selected downstream proxy tasks: depth completion, stereo matching, monocular feed-forward 3D scene reconstruction, SLAM, and vision-language spatial understanding. Unlike conventional evaluation protocols, our approach assesses DFMs based on their practical utility in real-world applications, bypassing problematic alignment procedures. 
We benchmark \textbf{\textit{eight}} state-of-the-art DFMs and provide an in-depth analysis of key findings and observations. We hope our work sparks further discussion in the community on best practices for depth model evaluation and paves the way for future research and advancements in depth estimation.

\end{abstract}

\newcommand{\starlegend}[2][1.0]{%
    \raisebox{-0.5ex}{ 
    \scalebox{#1}{ 
        \begin{tikzpicture}
        \color[HTML]{#2}
            \draw[-, ultra thick] (-0.26, 0) -- (0.26, 0);
            \node at (0, 0) {\small$\bigstar$};
        \end{tikzpicture}
    }}%
}

\section{Introduction}
\label{sec:intro}

Depth estimation plays a crucial role in various computer vision applications, from 3D scene reconstruction autonomous driving, to robotics  ~\cite{zhang2023controlnet,li2023bevdepth,zhu2024nicerslam,szymanowicz2024flash3d}. In recent years, deep learning-based approaches have significantly advanced the field, leading to powerful foundation models capable of generating high-quality depth predictions across diverse input domains~\cite{eigen2014mde,bhat2023zoedepth,ke2024repurposing,yang2024depthanythingv2,Ranftl2022midas,wang2024moge,wang2025vggt}. However, despite these advancements, evaluating and comparing depth estimation models remains an open challenge~\cite{ge2024geobench}. Existing evaluation protocols often overlook critical factors that impact both the validity and comparability of results.

A major limitation in current depth evaluation lies in its reliance on \textit{alignment-based metrics}, where predictions are aligned to ground truth before computing metrics. However, this alignment process introduces several biases that can affect fairness. Depth estimation methods adopt different representations—some predict metric depth directly, while others estimate affine-invariant disparity or affine-invariant depth, requiring distinct alignment strategies. Applying the same alignment solver across these representations can be problematic, as depth and disparity are related by a non-linear transformation. Furthermore, the widely used least squares solver is highly sensitive to outliers, favoring smoother depth predictions with lower global error while penalizing sharper estimates that may contain large gradients. These biases raise concerns about the robustness and fairness of existing evaluation protocols, as discussed in Sec.~\ref{sec:analysis}.

\begin{figure}[t]
    \centering
    \begin{tabular}{c|c}
        \multirow{2}{*}{
        \vtop{\null\hbox{\includegraphics[width=0.4\linewidth]{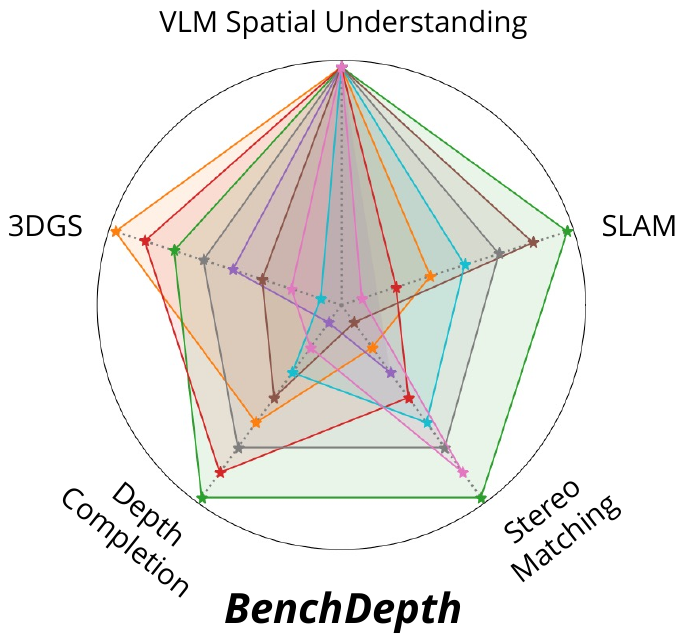}}}
        }
    
        &

        \vtop{\null\hbox{\includegraphics[width=0.3\linewidth]{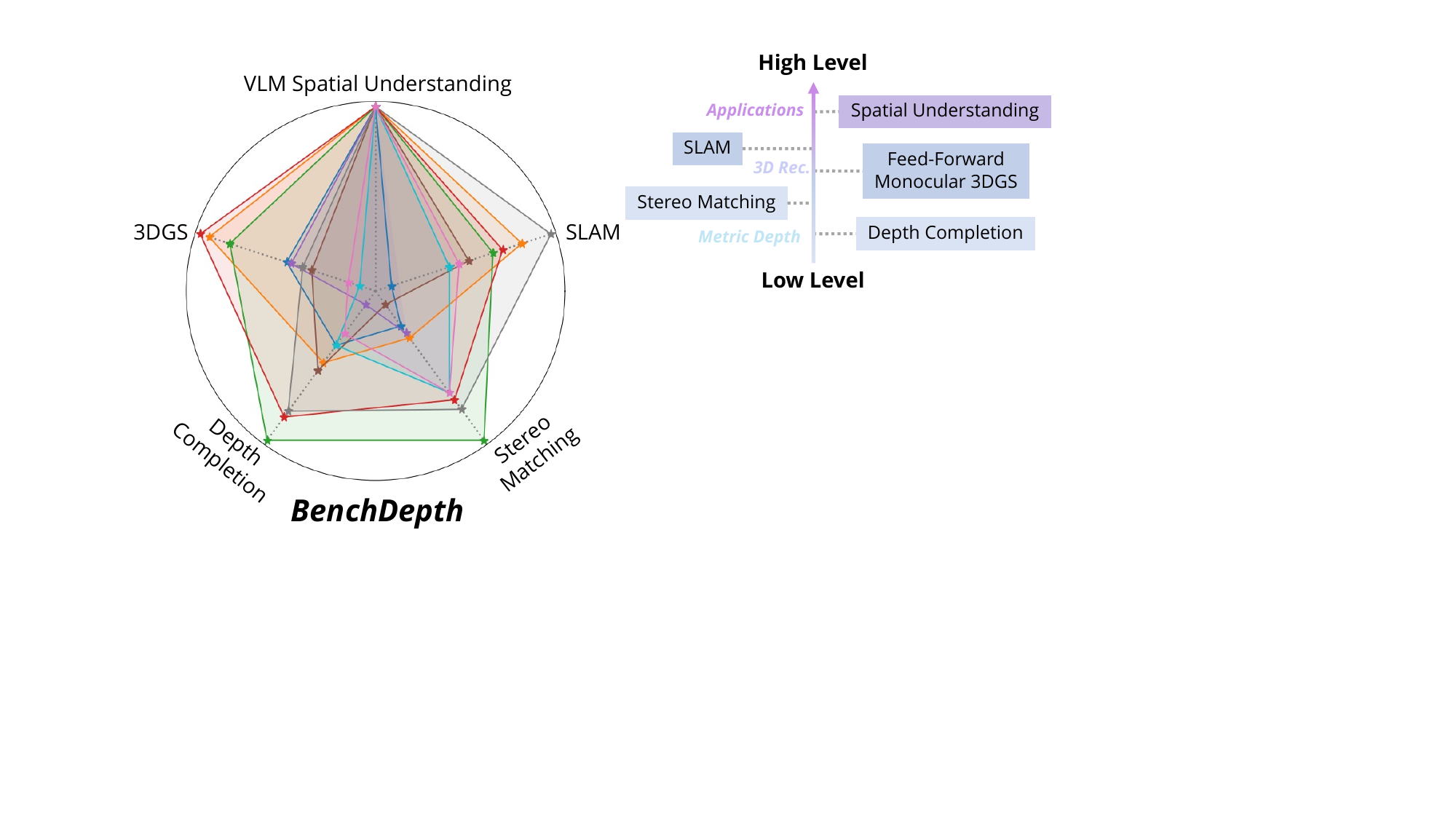}}}
        
        \\
        &
        \vspace{0.8cm}
        
        \vtop{\null\hbox{
            \scalebox{0.5}{
		\begin{tabular}[b]{L{1.2cm}|L{2.8cm}|C{2.6cm}|C{2.2cm}|C{1.4cm}}
        \hline
		Legend	&  Method & Type & Alignment & Rank$\downarrow$ \\
            \hline
    	\starlegend[0.8]{ff7f0e} & \cellcolor{Salmon}Midas~\cite{Ranftl2022midas} & \multirow{2}{*}{affine-inv. disp} & \multirow{2}{*}{(pred, 1/gt)} & 4.25 \\
            \starlegend[0.8]{2ca02c} & \cellcolor{Salmon}DAV2-Rel~\cite{yang2024depthanythingv2} & & & \textbf{1.50} \\
            \hline
            \starlegend[0.8]{d62728}&\cellcolor{Goldenrod}DAV2-Met~\cite{piccinelli2024unidepth} & \multirow{3}{*}{metric depth} & \multirow{3}{*}{w/o or (pred, gt)} &  \underline{3.75} \\
            \starlegend[0.8]{9467bd}&\cellcolor{Goldenrod}Metric3DV2~\cite{hu2024metric3d} & & & 6.33 \\
            \starlegend[0.8]{8c564b}&\cellcolor{Goldenrod}UniDepth~\cite{piccinelli2024unidepth} & & & 5.25 \\
            \hline
            \starlegend[0.8]{17becf}&\cellcolor{Apricot}Marigold~\cite{ke2024repurposing} & \multirow{2}{*}{affine-inv. depth} & \multirow{2}{*}{(pred, gt)} & 5.50 \\
            \starlegend[0.8]{7f7f7f}&\cellcolor{Apricot}GenPercept~\cite{xu2024genpercept} & & & 3.25 \\
            \hline
            \starlegend[0.8]{e377c2}&\cellcolor{LimeGreen}MoGe~\cite{wang2024moge} & affine-inv. pc & Various & 5.75 \\
            \hline
	\end{tabular}}

        }}
    \end{tabular}

    \caption{\textbf{BenchDepth illustration and results.} We evaluate different types of depth predictions (highlighted with different colors) with proxy tasks in a bottom-to-top manner, where MoGe~\cite{wang2024moge} adopts various alignment algorithms to compare with different types of depth methods. We show the rank of existing methods for each task on the left and present the average rank in the right table. Note that there are metric depth models, like Metric3DV2~\cite{hu2024metric3d}, using scale and shift during inference, conflicting with the task definition.}
    \label{fig:teaser}
\end{figure}


Additionally, inconsistencies arise when evaluating metric depth predictions. While some works directly compare metric depth to ground-truth values~\cite{piccinelli2024unidepth,hu2024metric3d}, others treat these predictions as scale- or affine-invariant and apply alignment before computing metrics~\cite{wang2024moge,ge2024geobench}. This lack of standardization complicates the interpretation of results and creates unnecessary variation in evaluation procedures. 

These challenges not only hinder fair comparisons among existing methods but also discourage the adoption of novel depth representations. For instance, MoGe~\cite{wang2024moge} estimates an affine-invariant point map, where z-coordinates represent affine-invariant depth. To compare MoGe with affine-invariant disparity-based methods, the authors must first recover the shift term via point map optimization, then convert it into scale-invariant depth, and finally invert it to obtain affine-invariant disparity before aligning it with the inverse of the ground-truth depth. This multi-step process complicates model evaluation and introduces additional sources of error due to multiple optimization steps.

Moreover, downstream tasks increasingly rely on depth as a guidance, emphasizing the need for an evaluation framework that can reveal a model’s potential across various applications~\cite{park2024depthprompt,szymanowicz2024flash3d,zhu2024nicerslam,jiang2025defom,cheng2025monster}. Traditional benchmarks primarily focus on numerical accuracy within constrained depth estimation settings, failing to assess how well different models generalize to real-world tasks~\cite{ge2024geobench}. 

To address these challenges, we propose a new approach for benchmarking depth foundation models. Rather than relying solely on traditional depth evaluation metrics, we use downstream tasks as proxy tasks for model evaluation. This approach is inspired by the success of large language model (LLM), vision language model (VLM), and image classification~\cite{achiam2023gpt,li2023blip,he2020moco,oquab2023dinov2}, where the evaluation is often based on downstream tasks. To this end, we propose \textbf{\textit{BenchDepth}}, a benchmark consisting of five downstream proxy tasks: stereo matching~\cite{xu2023igev}, depth completion~\cite{park2024depthprompt}, monocular feed-forward 3D scene reconstruction~\cite{szymanowicz2024flash3d}, SLAM~\cite{zhu2024nicerslam}, and 3D-VQA~\cite{zuo2024towards}.  The tasks are selected in a bottom-to-top manner as shown in Fig.~\ref{fig:teaser}, ranging from applications in low-level to high-level vision.
These tasks allow us to evaluate the practical utility of depth foundation models in a fair manner, without relying on potentially problematic depth alignment procedures.

In this paper, we fairly benchmark \textbf{\textit{eight}} state-of-the-art 3D foundation models with DepthBench. By examining their performance on a set of proxy tasks, we provide a more robust and holistic understanding of what constitutes a good foundation depth model. Our main findings and conclusions are as follows:

\begin{enumerate}
    \item Most depth foundation models improve the performance of downstream tasks, highlighting their potential for broader applications in the future.
    \item Overall, DAV2~\cite{yang2024depthanythingv2} achieves the best results across proxy tasks, demonstrating the benefits of scaling up training data and incorporating synthetic data.
    \item Affine-invariant disparity methods consistently outperform other depth estimation approaches, even with MiDaS~\cite{Ranftl2022midas} being the oldest method among them.
    \item Despite being fine-tuned on a single dataset (Hypersim~\cite{roberts2021hypersim}, synthetic), DAV2-Met significantly outperforms other metric depth models~\cite{hu2024metric3d,piccinelli2024unidepth} trained on multiple datasets. This aligns with the conclusion of ZoeDepth~\cite{bhat2023zoedepth} that fine-tuning a well-pretrained affine-invariant disparity model enhances metric depth estimation. Moreover, the performance gap suggests that incorporating synthetic data for metric depth training is crucial, as it allows models to learn high-frequency details that are often lost in real-world datasets~\cite{yang2024depthanythingv2,li2024patchrefiner}.
    \item The performance improvement from Marigold~\cite{ke2024repurposing} to GenPercept~\cite{xu2024genpercept} underscores the importance of effective fine-tuning strategies for Stable Diffusion~\cite{rombach2022sd}, a powerful foundation model. Expanding the training data could further unlock their potential, following the success of other methods, as the current fine-tuning process is limited to VKITTI~\cite{geiger2013vkitti} and Hypersim.
    \item MoGe~\cite{wang2024moge}, as a novel approach for geometry estimation, demonstrates potential on DepthBench, though further research is needed to improve its performance.
    \item For the highest-level task, VLM spatial understanding, all methods yield comparable results. This suggests that at this higher level, different depth estimation approaches can be equally effective.
\end{enumerate}

We hope that our work will spark further discussion in the community about the best practices for depth model evaluation and pave a way for the further research and development of depth estimation.



\section{Related Works}
\label{sec:related}

\subsection{Depth Foundation Model (DFM)}
Monocular depth estimation has seen significant advancements with the availability of large-scale public datasets~\cite{silberman2012nyu,geiger2012kitti,cordts2016cityscapes}, improved architectural designs~\cite{eigen2014mde,li2023depthformer,bhat2021adabins,li2023patchfusion}, and enhanced training strategies~\cite{chen2016diw,fu2018dorn,li2022binsformer}, \etc. While earlier works primarily focused on achieving high performance in in-domain inference, the scaling of both models and datasets in deep learning~\cite{kaplan2020scaling} has shifted recent research toward developing foundation models with strong zero-shot generalization across unseen domains (\ie, diverse real-world images). For example, MiDaS~\cite{Ranftl2022midas} introduces a mixture-dataset training approach and adopted an affine-invariant disparity representation to handle cross-dataset inconsistencies. DAV2~\cite{yang2024depthanything,yang2024depthanythingv2} follows a similar formulation but scaled training further using a semi-supervised learning paradigm. Other works  leverage the prior knowledge of Stable Diffusion~\cite{rombach2022sd} and fine-tune pretrained models for affine-invariant depth estimation~\cite{ke2024repurposing,xu2024genpercept}. Other lines of research such as Metric3DV2~\cite{hu2024metric3d} and UniDepth~\cite{piccinelli2024unidepth} aim to predict metric depth by incorporating explicit camera models. MoGe~\cite{wang2024moge} proposes a novel formulation using affine-invariant point maps~\cite{wang2024dust3r} to represent monocular geometry. Despite the rapid progress in depth foundation models, a key challenge remains: fairly evaluating and comparing these models across different depth representations and real-world applications.

\subsection{Evaluations of DFMs}

Eigen \etal~\cite{eigen2014mde} introducs the first deep learning framework for monocular \textit{metric} depth estimation, along with several standard evaluation metrics that remain widely used today. However, while depth estimation methods have diversified into various depth representations (as summarized in Tab.~\ref{fig:teaser}), existing works attempt to adopt the same evaluation protocol designed for metric depth estimation~\cite{Ranftl2022midas,yang2024depthanythingv2,hu2024metric3d,ke2024repurposing,xu2024genpercept,wang2024moge}. This might lead to seemingly comparable numerical results that may still be biased due to inconsistencies in the alignment process. Inspired by large language models (LLMs)~\cite{achiam2023gpt}, vision-language models (VLMs)~\cite{li2023blip}, and self-supervised learning in image classification~\cite{he2020moco,oquab2023dinov2}, where the evaluation is often based on downstream tasks, we propose a proxy-task-based benchmark for DFMs. By assessing DFMs on a diverse set of real-world tasks in a bottom-to-top manner, our approach enables a fairer and more practical comparison, eliminating the need for problematic alignment procedures. Compared with \cite{cong2025e3dbench}, our benchmark focuses on the monocular setting and the practical potential for downstream tasks.

\SetKwComment{Comment}{/* }{ */}

\begin{algorithm}
\caption{Robustness analysis of the alignment algorithm in depth and disparity spaces.}\label{alg:robust}
\KwData{Matrix size $n = 500$; Max disturbance factor $m = 1.8$; Alignment flag $align = true$}
\KwResult{Computed metrics: $\delta_1$ and $\delta_2$}
\Comment{Initialize matrices}

$\boldsymbol{GT}_{depth}$ $\gets$ a random $(n, n)$ matrix $\in[0, 10]$\;
$\boldsymbol{GT}_{disparity}$ $\gets$ 1 / $\boldsymbol{GT}_{depth}$\;
$\boldsymbol{Pred}_{depth}$ $\gets$ $\boldsymbol{GT}_{depth}$ \;
$\boldsymbol{Pred}_{disparity}$ $\gets$ $\boldsymbol{GT}_{disparity}$ \;

\Comment{Increasing disturbance}
Define \texttt{dist} as sequence $[0, 0.05, \dots, m]$\;

\ForEach{$d$ in \texttt{dist}}{
    Generate a random $(n, n)$ error matrix from a Gaussian distribution $\mathcal{N}(0, d \times 0.01)$: $\boldsymbol{E}$\;

    \Comment{Apply the disturbance}
    $\boldsymbol{Pred'}_{depth}$ = $\boldsymbol{Pred}_{depth} + \boldsymbol{E}$\;
    $\boldsymbol{Pred'}_{disparity}$ = $\boldsymbol{Pred}_{disparity} + \boldsymbol{E}$\;
    Compute metrics with alignments in depth and disparity spaces, respectively\;
    \quad $\delta_1$ $\gets \text{metric($\boldsymbol{GT}$, $\boldsymbol{Pred'}_{depth}$, $align$)}$\;
    \quad $\delta_2$ $\gets \text{metric($\boldsymbol{GT}$, $\boldsymbol{Pred'}_{disparity}$, $align$)}$\;
}

\end{algorithm}


\section{Overlooked Alignment for Evaluating DFMs}
\label{sec:analysis}

In this section, we analyze the limitations of the current depth evaluation protocols. We focus on the widely used metric, $\delta$, which measures the proportion of pixels satisfying $\max(a_i/d_i, d_i/a_i) < 1.25$, where $a$ and $d$ are the aligned prediction and ground-truth depth, respectively.

\subsection{Alignment in Different Spaces}

As summarized in Tab.~\ref{fig:teaser}, affine-invariant disparity estimation methods align their predictions with the inverse of the ground-truth depth, while other methods align predictions directly in depth space. The commonly used least squares solver for alignment is designed for ordinary \textit{linear} first-order differential equations, but the inverse operator is inherently \textit{non-linear}. This discrepancy introduces different behaviors when aligning predictions in these two spaces, leading to potential unfairness in comparisons.

To analyze the robustness of the alignment process in different spaces, we conduct an experiment (Alg.~\ref{alg:robust}) where a magnifying disturbance is added to the predicted depth and disparity, both initialized as ground-truth values. The standard protocol is then applied to compute the evaluation metric after alignment. As illustrated in Fig.~\ref{fig:algo_1}, aligning in disparity space exhibits higher robustness to small errors compared to depth space. However, it becomes more sensitive to larger errors. This asymmetric behavior in different alignment spaces introduces inconsistencies in evaluation, revealing issues in current depth evaluation protocols.


\subsection{Sensitivity of Scale-and-Shift Alignment}

Since the least squares solver is sensitive to large outliers~\cite{lawson1995solving,heath2018scientific,wang2024moge}, we conduct an experiment to investigate its impact on the depth evaluation metric. In Alg.~\ref{alg:metrics}, we initialize a predicted depth map identical to the ground-truth depth and introduce a disturbance with decreasing size. We then compute depth metrics with and without alignment.

\begin{figure}[t]
  \centering
  \begin{subfigure}{0.4\textwidth}
    \centering
    \includegraphics[width=\textwidth]{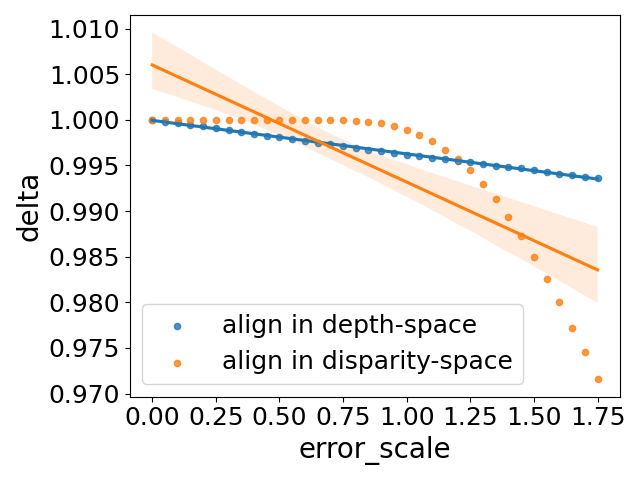}
      \caption{Algo.~\ref{alg:robust} plot.}
      \label{fig:algo_1}
  \end{subfigure}
  \hfill
  \begin{subfigure}{0.4\textwidth}
    \centering
    \includegraphics[width=\textwidth]{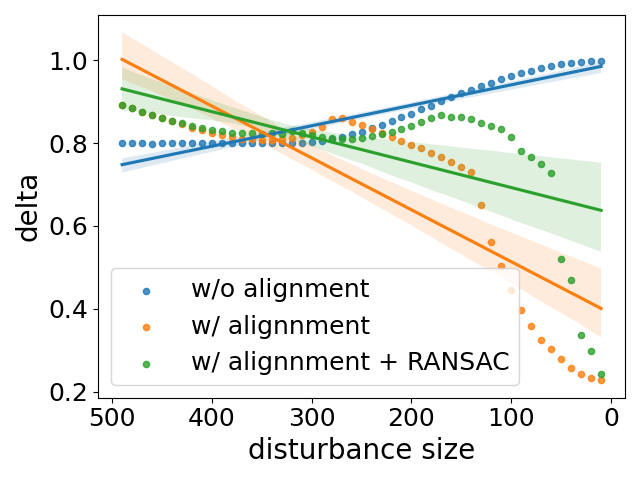}
      \caption{Algo.~\ref{alg:metrics} plot.}
      \label{fig:algo_2}
  \end{subfigure}
  \caption{\textbf{(a)} Aligning in the disparity space exhibits higher robustness to small errors compared to depth space. However, it becomes more sensitive to larger errors. \textbf{(b)} The presence of outliers can significantly disrupts the alignment results, leading to an entirely different monotonicity patterns for the same metric with and without alignment.}
  \label{fig:algos}
\end{figure}

Fig.~\ref{fig:algo_2} reveals that the $\delta$ metric exhibits entirely different monotonicity patterns depending on whether alignment is applied. Without alignment, the metric behaves as expected: as the disturbance size decreases, the proportion of pixels satisfying the accuracy threshold increases. However, with alignment, the presence of outliers significantly disrupts the alignment results, leading to a counterintuitive $\delta$ metric that fails to accurately reflect depth prediction quality. Adopting RANSAC to filter outliers can alleviate the impact, but the issue still exists. This suggests that alignment biases the evaluation protocol in favor of smoother depth predictions, while sharper depth maps, which can introduce stronger outlier gradients, suffer from degraded evaluation scores.

To eliminate biases in alignment, we propose benchmarking depth estimation using proxy tasks. By directly feeding depth predictions into proxy task frameworks without any alignment, we enable a fair comparison among different depth estimation methods, independent of scale and shift variations.

\SetKwComment{Comment}{/* }{ */}

\begin{algorithm}[t!]
\caption{Analysis of the influence of the alignment algorithm's sensitivity to the depth metric.}\label{alg:metrics}
\KwData{Matrix size $n = 500$; Alignment flag $align = true/false$}
\KwResult{Computed metrics: $\delta$, AbsRel with local disturbance}
\Comment{Initialize matrices}
$\boldsymbol{Pred}$ $\gets$ a random $(n, n)$ matrix $\in[0, 10]$\;
$\boldsymbol{GT}$ $\gets$ $\boldsymbol{Pred}$\;

\Comment{Local disturbance}
Define \texttt{size} as sequence $[n, \dots, 20, 10]$\;

\ForEach{$m$ in \texttt{size}}{
    \Comment{Make a copy}
    $\boldsymbol{Pred}_c$ $\gets \boldsymbol{Pred}$\;
    Init a random $(m, m)$ error matrix: $\boldsymbol{E} \in [0, 1]$\;
    \Comment{Modify the top-left region}
    \texttt{$\boldsymbol{Pred_c}$[:$m$, :$m$] += $\boldsymbol{E}$ *} $\frac{n^2}{m^2}$\;
    Compute metrics:\;
    \quad $\delta$ $\gets \text{metric($\boldsymbol{GT}$, $\boldsymbol{Pred_c}$, $align$)}$\;
}

\end{algorithm}

\section{BenchDepth}
\label{sec:benchmark}

We introduce \textbf{BenchDepth}, a novel benchmark for depth estimation, designed with carefully selected proxy tasks in a bottom-up manner (Fig.~\ref{fig:teaser}). As lower-level tasks, we select depth completion~\cite{park2024depthprompt} and stereo matching~\cite{xu2023igev}. These tasks closely resemble depth estimation, as they belong to the category of metric depth estimation but incorporate additional prompts (\eg, sparse depth from real sensors or stereo image pairs with a fixed baseline). Middle-level tasks feed-forward 3D Gaussian Splatting (3DGS)~\cite{szymanowicz2024flash3d} and SLAM~\cite{zhu2024nicerslam}, focus on 3D reconstruction but differ in representation (3DGS~\cite{kerbl20233dgs} and neural implicit representations~\cite{xie2022neural}) and the number of input images (single or multiple). At the highest level, we evaluate depth estimation for vision-language models (VLMs)~\cite{cai2024spatialbot}, aiming to assess the role of depth in enhancing spatial understanding. 

Selected depth foundation estimation methods for benchmarking are summarized in Tab.~\ref{fig:teaser}. We choose the most representative methods from each depth estimation category. Note that though DAV2-Met~\cite{yang2024depthanythingv2}, Metric3DV2~\cite{hu2024metric3d}, and UniDepth~\cite{piccinelli2024unidepth} are all metric methods, DAV2-Met is fine-tuned on a single metric dataset (Hypersim~\cite{roberts2021hypersim}), whereas the other two methods are trained with a mixture of many datasets. We use the default camera parameter assumption for Metric3DV2 and UniDepth. Since the original version of Marigold~\cite{ke2024repurposing} is hard to be adopted to online training due to the large number of inference steps, we use the end-to-end fine-tuned version of Marigold~\cite{martingarcia2024diffusione2eft} that supports one-step inference as a replacement.

Key features of BenchDepth include:
\begin{enumerate}
    \item Fair comparisons among various DFMs without reliance on alignment.
    \item Evaluation of the broader applicability of DFMs beyond standard benchmarks~\cite{ge2024geobench}.
\end{enumerate}

Below, present the five proxy tasks in detail and describe the modifications applied to selected methods to support depth evaluation using DepthBench. We use 8 GPUs to conduct the benchmark.

\begin{figure}[t]
\captionsetup[subfigure]{labelformat=empty}
  \centering
  \begin{subfigure}{0.45\textwidth}
    \centering
    \includegraphics[width=\textwidth]{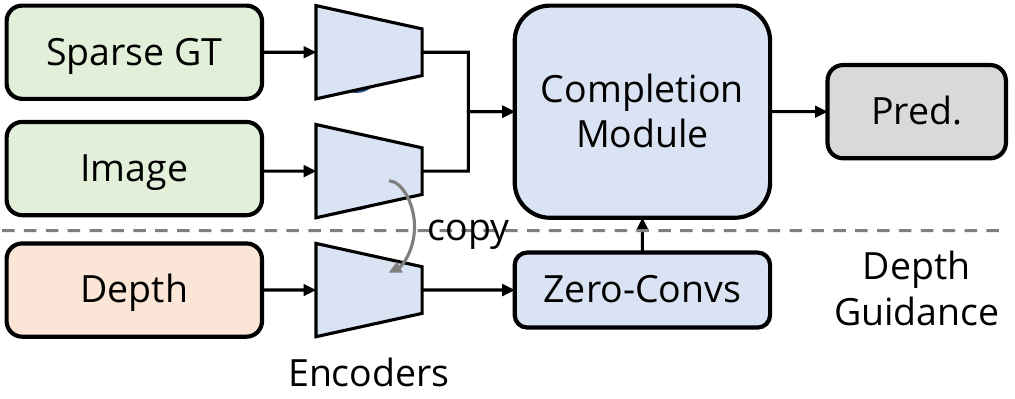}
      \caption{(a)}
      \label{fig:completion}
  \end{subfigure}
  \hfill                              
  \begin{subfigure}{0.45\textwidth}
    \centering
    \includegraphics[width=\textwidth]{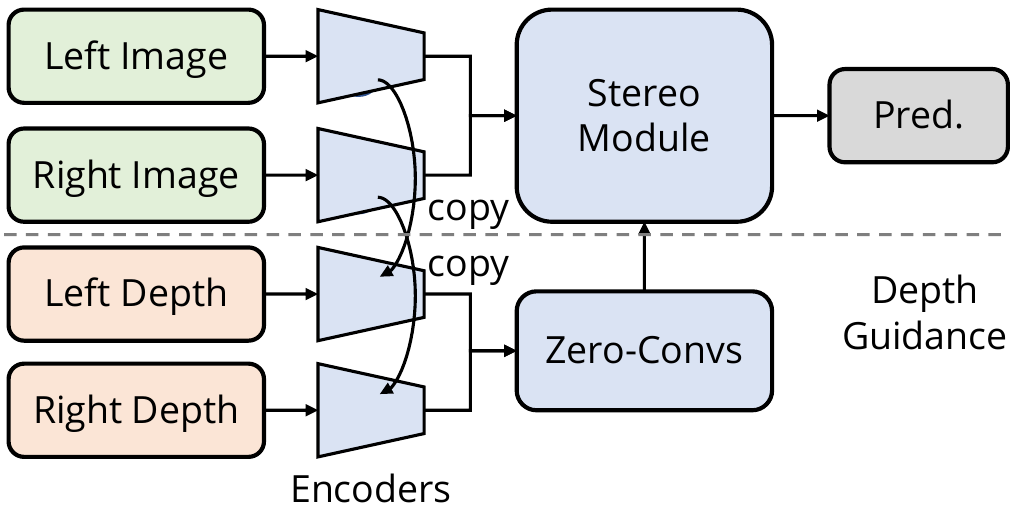}
      \caption{(b)}
      \label{fig:stereo}
  \end{subfigure}
  \caption{\textbf{(a)} Depth completion framework and \textbf{(b)} Stereo matching framework for depth benchmark. We adopt zero convolutions~\cite{zhang2023controlnet} to introduce depth guidance without modifying core components of proxy tasks.}
  \label{fig:framework}
\end{figure}

\begin{table*}[t!]
    \centering
    \caption{\textbf{Benchmark with metric depth completion.} We select DepthPrompting~\cite{park2024depthprompt} as the baseline method and apply depth predictions from various foundation models as the guidance. We use different amounts of sparse samples (from 100 to 1) in this experiment. Best results are in \textbf{bold}, second best are \underline{underlined}. \textit{imp. (\%)} indicates the average improvement ratio, and \textit{rank} is calculated based on it.}
    \label{tab:benchmark_prompting}
    \scalebox{0.72}{
    \begin{tabular}{L{2.4cm}|*{2}{C{0.8cm}}|*{2}{C{0.8cm}}|*{2}{C{0.8cm}}|*{2}{C{0.8cm}}|*{2}{C{0.8cm}}|C{0.9cm}|C{0.9cm}}
        \toprule
        \multirow{2}{*}{Method} & \multicolumn{2}{c|}{100} & \multicolumn{2}{c|}{32} & \multicolumn{2}{c|}{8} & \multicolumn{2}{c|}{4} & \multicolumn{2}{c|}{1} & \multirow{2}{*}{\textit{\textit{imp.}}} & \multirow{2}{*}{\textit{rank}}\\
        & RMSE & MAE & RMSE & MAE & RMSE & MAE & RMSE & MAE  & RMSE & MAE & & \\
        \midrule
        w/o depth~\cite{park2024depthprompt}&0.206&\underline{0.102}&0.334&0.199&0.486&0.340&0.514&0.370&0.550&0.406&-&- \\
        \cellcolor{Salmon}Midas~\cite{Ranftl2022midas}&0.204&0.114&0.294&0.182&0.449&0.311&0.493&0.355&0.556&0.414&\textcolor{PineGreen}{+3.09}&4\\
        \cellcolor{Salmon}DAV2-Rel~\cite{yang2024depthanythingv2} &\textbf{0.191}&\textbf{0.099}&\textbf{0.279}&\textbf{0.166}&\textbf{0.427}&\textbf{0.292}&\textbf{0.471}&\textbf{0.336}&\underline{0.533}&\underline{0.396}&\textcolor{PineGreen}{+\textbf{9.26}}&\textbf{1}\\
        \cellcolor{Goldenrod}DAV2-Met~\cite{yang2024depthanythingv2}&0.202&0.112&0.287&0.178&\underline{0.431}&\underline{0.297}&\underline{0.472}&\underline{0.338}&\textbf{0.529}&\textbf{0.392}&\textcolor{PineGreen}{+6.48}&\underline{2}\\
        \cellcolor{Goldenrod}Metric3DV2~\cite{hu2024metric3d}&0.216&0.128&0.306&0.195&0.454&0.317&0.497&0.359&0.557&0.415&\textcolor{Maroon}{-0.38}&8 \\
        \cellcolor{Goldenrod}UniDepth~\cite{piccinelli2024unidepth}&0.210&0.122&0.296&0.187&0.438&0.308&0.480&0.349&0.540&0.404&\textcolor{PineGreen}{+2.97}&5\\
        \cellcolor{Apricot}Marigold~\cite{ke2024repurposing}&0.210&0.121&0.296&0.187&0.448&0.314&0.491&0.356&0.555&0.414&\textcolor{PineGreen}{+1.76}&6\\
        \cellcolor{Apricot}GenPercept~\cite{xu2024genpercept}&\underline{0.199}&0.110&\underline{0.284}&\underline{0.174}&0.436&0.301&0.479&0.342&0.542&0.402&\textcolor{PineGreen}{+\underline{6.16}}&3 \\
        \cellcolor{LimeGreen}MoGe~\cite{wang2024moge}&0.210&0.124&0.295&0.188&0.444&0.312&0.489&0.355&0.558&0.417&\textcolor{PineGreen}{+1.53}&7\\
        \bottomrule
    \end{tabular}
    }
\end{table*}

\textbf{Depth Completion:} Given sparse metric-scale depth measurements from sensors (\eg LiDAR, Radar) and corresponding images, depth completion aims to generate dense metric depth predictions. We select DepthPrompting~\cite{park2024depthprompt} as the baseline method. While DepthPrompting enables the adaptation of foundation depth models for completion, its reliance on feature extractors from these models~\cite{li2023depthformer} introduces bias, as the extractor quality may influence performance more than the predicted depth itself. To mitigate this, we standardize feature extractors across models and inject depth predictions using zero convolutions~\cite{zhang2023controlnet} (Fig.~\ref{fig:completion}). Additionally, we omit the alignment module in DepthPrompting to enable direct comparisons across depth methods. We use the NYU Depth V2 dataset~\cite{silberman2012nyu} for this proxy task, following the official split with about 50k training samples and 654 testing samples.

\textbf{Stereo Matching:} This task estimates disparity from two images with a known baseline. Metric depth can be recovered from disparity using camera parameters. We adopt IGEV~\cite{xu2023igev} as our baseline and incorporate zero convolutions~\cite{zhang2023controlnet} to inject depth predictions as shown in Fig.~\ref{fig:stereo}. Unlike prior works that develop task-specific strategies to integrate depth into stereo matching models~\cite{cheng2025monster,jiang2025defom}, our simple yet general approach allows for a more straightforward assessment of depth prediction quality. We use the SceneFlow dataset~\cite{mayer2016sceneflow}, which contains 35,454 training pairs and 4,370 test pairs with dense disparity maps. Middlebury 2014~\cite{scharstein2014middlebury} and ETH3D~\cite{schops2017eth3d} are used for zero-shot evaluation. 


\begin{table*}[t!]
    \centering
    \caption{\textbf{Benchmark with stereo matching.} We select IGEV~\cite{xu2023igev} as the baseline method and apply depth predictions from various foundation models as the guidance to fine-tune the baseline model. We present \textit{rank} for each each dataset whereas \textit{avg. rank} indicates the average rank of all evaluation performances.}
    \label{tab:benchmark_stereo_matching}
    \scalebox{0.73}{
    \begin{tabular}{L{2.4cm}|*{2}{C{1.6cm}}|*{2}{C{1.6cm}}|*{2}{C{1.6cm}}|C{1.2cm}|C{1.2cm}}
        \toprule
        \multirow{2}{*}{Method} & \multicolumn{2}{c|}{SceneFlow~\cite{mayer2016sceneflow}} & \multicolumn{2}{c|}{Middlebury~\cite{scharstein2014middlebury}} & \multicolumn{2}{c|}{ETH3D~\cite{schops2017eth3d}} & \multirow{2}{*}{\textit{imp.}} & \multirow{2}{*}{\textit{rank}} \\
        & EPE$\downarrow$ & $>$3pt$ (\%)\downarrow$ & EPE$\downarrow$ & $>$2pt$ (\%)\downarrow$ & EPE$\downarrow$ & $>$1pt$ (\%)\downarrow$ & &  \\
        \midrule
        w/o depth~\cite{xu2023igev} & 0.496 & 2.599 & \underline{0.857} & 6.655 & 0.283 & 3.575 & - & - \\
        \cellcolor{Salmon}Midas~\cite{Ranftl2022midas} & 0.483 & 2.502 & 1.061 & 7.316 & 0.273 & 3.383  & \textcolor{Maroon}{-3.07} & 7 \\
        \cellcolor{Salmon}DAV2-Rel~\cite{yang2024depthanythingv2} & \textbf{0.456} & \textbf{2.432} & \textbf{0.834} & 6.399 & 0.275 & \textbf{3.189}  & \textcolor{PineGreen}{+\textbf{5.77}} & \textbf{1} \\
        \cellcolor{Goldenrod}DAV2-Met~\cite{yang2024depthanythingv2} & \underline{0.471} & \underline{2.473} & 0.938 & \underline{6.177} & \underline{0.270} & 3.698 & \textcolor{PineGreen}{+1.46} & 5\\
        \cellcolor{Goldenrod}Metric3DV2~\cite{hu2024metric3d} & 0.482 & 2.521 & 0.949 & 7.309 & 0.275 & 3.523 & \textcolor{Maroon}{-1.74} & 6\\
        \cellcolor{Goldenrod}UniDepth~\cite{piccinelli2024unidepth} & 0.477 & 2.521 & 0.964 & 7.242 & 0.285 & 3.822 & \textcolor{Maroon}{-3.68} & 8\\
        \cellcolor{Apricot}Marigold~\cite{ke2024repurposing} & 0.475 & 2.499 & 0.899 & 6.519 & 0.273 & 3.485 & \textcolor{PineGreen}{+1.87} & 4 \\
        \cellcolor{Apricot}GenPercept~\cite{xu2024genpercept} & 0.473 & 2.485 & 0.935 & 6.649 &\textbf{0.265} & \underline{3.374} & \textcolor{PineGreen}{+1.99} & 3\\
        \cellcolor{LimeGreen}MoGe~\cite{wang2024moge} & 0.473 & 2.481 & 0.907 & \textbf{5.951}  & 0.279 & 3.544 & \textcolor{PineGreen}{+\underline{2.70}} & \underline{2}\\
        \bottomrule
    \end{tabular}
    }
\end{table*}

\begin{table*}[t!]

    \centering
    \caption{\textbf{Benchmark with feed-forward monocular 3D scene reconstruction by novel view synthesis.} We select Flash3D~\cite{szymanowicz2024flash3d} as the baseline method and apply depth predictions from various foundation models as the model input. Following \cite{szymanowicz2024flash3d}, we present results of small, medium and large baseline ranges separately.}
    \label{tab:benchmark_ffgs}
    \scalebox{0.73}{
    \begin{tabular}{L{2.4cm}|*{3}{C{1.0cm}}|*{3}{C{1.0cm}}|*{3}{C{1.0cm}}|C{1.2cm}|C{1.2cm}}
        \toprule
        \multirow{2}{*}{Method} & \multicolumn{3}{c|}{5 frames} & \multicolumn{3}{c|}{10 frames} & \multicolumn{3}{c|}{$\mathcal{U}\left[-30, 30\right]$ frames} & \multirow{2}{*}{\textit{imp}} & \multirow{2}{*}{\textit{rank}}\\
        & PSNR$\uparrow$ & SSIM$\uparrow$ & LPIP$\downarrow$ & PSNR$\uparrow$ & SSIM$\uparrow$ & LPIP$\downarrow$ & PSNR$\uparrow$ & SSIM$\uparrow$ & LPIP$\downarrow$ & & \\
        \midrule
        w/o depth~\cite{szymanowicz2024flash3d} & 24.285&\underline{0.803}&0.151&21.767&0.729&0.203&21.241&0.705&0.230 & & \\
        \cellcolor{Salmon}Midas~\cite{Ranftl2022midas} &24.964&\textbf{0.812}&\textbf{0.125}&22.290&\textbf{0.735}&\textbf{0.179}&\underline{21.769}&\underline{0.710}&\textbf{0.212}&\textcolor{PineGreen}{+\textbf{5.24}}&\textbf{1} \\
        \cellcolor{Salmon}DAV2-Rel~\cite{yang2024depthanythingv2}&\underline{24.965}&\textbf{0.812}&0.129&\underline{22.305}&\underline{0.733}&0.185&21.703&0.706&0.218&\textcolor{PineGreen}{+4.21}&3 \\
        \cellcolor{Goldenrod}DAV2-Met~\cite{yang2024depthanythingv2} & \textbf{25.000}&\textbf{0.812}&\underline{0.128}&\textbf{22.341}&\textbf{0.735}&\underline{0.182}&\textbf{21.842}&\textbf{0.711}&\underline{0.215}&\textcolor{PineGreen}{+\underline{4.81}}&\underline{2} \\
        \cellcolor{Goldenrod}Metric3DV2~\cite{hu2024metric3d} &24.468&0.787&0.150&21.994&0.713&0.204&21.396&0.690&0.233&\textcolor{Maroon}{-0.05}&5\\
        \cellcolor{Goldenrod}UniDepth~\cite{piccinelli2024unidepth}&23.983&0.786&0.145&21.530&0.708&0.202&21.036&0.687&0.235&\textcolor{Maroon}{-0.10}&6\\
        \cellcolor{Apricot}Marigold~\cite{ke2024repurposing} &23.974&0.779&0.162&21.515&0.701&0.219&20.952&0.676&0.248&\textcolor{Maroon}{-4.19}&8\\
        \cellcolor{Apricot}GenPercept~\cite{xu2024genpercept} &24.119&0.787&0.140&21.489&0.705&0.197&21.029&0.682&0.230&\textcolor{Maroon}{-0.14}&4\\
        \cellcolor{LimeGreen}MoGe~\cite{wang2024moge} & 23.930&0.780&0.144&21.309&0.696&0.202&20.851&0.673&0.235&\textcolor{Maroon}{-1.60}&7\\
        \bottomrule
    \end{tabular}
    }
    
\end{table*}

\textbf{Feed-Forward Monocular 3DGS:} This task reconstructs scenes and synthesizes novel views from a single image using 3D Gaussian Splatting~\cite{kerbl20233dgs}. We use Flash3D~\cite{szymanowicz2024flash3d} as the baseline model. Flash3D incorporates a frozen depth foundation model in its first stage to estimate depth from the input image. The predicted depth and image are then processed by a UNet-like~\cite{ronneberger2015unet} network to estimate 3DGS parameters. Since the foundation depth model remains frozen and no features from the foundation model are used in the second stage, we can adopt different foundation models for the first stage and train Flash3D following the default recipe. We use the RealEstate10k dataset~\cite{zhou2018stereo}. It consists of real estate videos from YouTube, with 67,477 training scenes and 7,289 test scenes. Some outdated samples were removed, causing slight deviations from the results reported in~\cite{szymanowicz2024flash3d}.

\textbf{Simultaneous Localization and Mapping:} Simultaneous Localization and Mapping (SLAM) is a fundamental problem in computer vision with broad applications. We employ NICER-SLAM~\cite{zhu2024nicerslam} as our baseline, as it integrates dense SLAM with a neural implicit representation for tracking and mapping from monocular RGB videos. Since NICER-SLAM can process RGB-D sequences, we replace the original sensor depth with depth predictions from different foundation models and train the system accordingly. To better assess the impact of depth predictions, we omit pseudo-depth loss during training. We evaluate models on the Replica dataset~\cite{straub2019replica}, which provides RGB-(D) images rendered using the official renderer. All 8 scenes are used for benchmarking. For benchmarking, we replace the original input depth with estimated depth from different methods and omit the monocular depth loss (Eq.~13 in~\cite{zhu2024nicerslam}), which depends on another depth model. We exclude Metric3DV2 since it was trained on this dataset, though there is no evidence of overfitting. 


\textbf{VLM Spatial Understanding:} Vision-Language Models (VLMs) have demonstrated strong performance in 2D image understanding but remain limited in spatial reasoning~\cite{cai2024spatialbot}. Since depth maps contain spatial information, incorporating them as additional inputs may improve VLMs' 3D understanding. For this proxy task, we adopt SpatialBench~\cite{cai2024spatialbot} to evaluate the impact of different depth models on VLM spatial reasoning. We use two VLMs: ChatGPT-4o and SpatialBot-Phi2-3B~\cite{cai2024spatialbot}. Since ChatGPT-4o is not trained with depth maps, we render depth predictions using the magma colormap and provide corresponding text prompts. 





\section{Benchmark Results}
\label{sec:exp}

\noindent\textbf{Depth Completion.} Tab.~\ref{tab:benchmark_prompting} presents the benchmark results. DAV2-Rel~\cite{yang2024depthanythingv2} is the only method that consistently improves performance across almost all settings, achieving rank 1. Most methods provide a performance boost, except for Metric3DV2~\cite{hu2024metric3d}, which performs worse that the baseline. Interestingly, depth methods tend to be more beneficial when the available sparse ground-truth (GT) depth is limited. This suggests that foundation models provide useful guidance when GT depth is scarce. However, as GT depth increases, the ambiguity in selecting the appropriate depth source limits further improvements compared to using only sparse GT depth for guidance.

\noindent\textbf{Stereo Matching.} Tab.~\ref{tab:benchmark_stereo_matching} presents the results for stereo matching. In the in-domain setting, all foundation depth models significantly improve baseline performance, with an average 4.5\% EPE gain. However, in zero-shot cross-domain evaluation, not all methods generalize well. DAV2-Rel, GenPercept~\cite{xu2024genpercept}, and Marigold~\cite{ke2024repurposing} perform best. Metric depth models, such as Metric3DV2~\cite{hu2024metric3d} and UniDepth~\cite{piccinelli2024unidepth}, underperform compared to other types of depth estimation methods. Notably, DAV2-Met~\cite{yang2024depthanythingv2} outperforms other metric depth models, possibly benefiting from fine-tuning DAV2-Rel, despite being trained on only one dataset (Hypersim~\cite{roberts2021hypersim}). The ability of DAV2-Met to predict sharper metric depth may also contribute to its superior performance.

\begin{table*}[t]
    \centering
    \caption{\textbf{Benchmark with Simultaneous Localization and Mapping (SLAM).} We select Nicer-SLAM~\cite{zhu2024nicerslam} as the baseline method and apply depth predictions from various foundation models as the model input. acc and com are short for accuracy and completion, respectively. Rendered indicates that the input depth map is rendered by the dataset. We exclude Metric3DV2 and use \textcolor{gray}{\textbf{gray}} for its results as it is trained with this dataset.}
    \label{tab:benchmark_slam}
    \scalebox{0.59}{
    \begin{tabular}{L{2.4cm}|*{2}{C{0.7cm}}|*{2}{C{0.7cm}}|*{2}{C{0.7cm}}|*{2}{C{0.7cm}}|*{2}{C{0.7cm}}|*{2}{C{0.7cm}}|*{2}{C{0.7cm}}|*{2}{C{0.7cm}}|C{0.8cm}C{0.5cm}}
        \toprule
        \multirow{2}{*}{Method} & \multicolumn{2}{c|}{rm-0} & \multicolumn{2}{c|}{rm-1} & \multicolumn{2}{c|}{rm-2} & \multicolumn{2}{c|}{off-0} & \multicolumn{2}{c|}{off-1} & \multicolumn{2}{c|}{off-2} & \multicolumn{2}{c|}{off-3} & \multicolumn{2}{c|}{off-4} & 
        \multirow{2}{*}{\textit{imp.}} & \multirow{2}{*}{\textit{rank}} \\
        & acc$\downarrow$ & com$\downarrow$ & acc$\downarrow$ & com$\downarrow$ & acc$\downarrow$ & com$\downarrow$ & acc$\downarrow$ & com$\downarrow$ & acc$\downarrow$ & com$\downarrow$  & acc$\downarrow$ & com$\downarrow$ & acc$\downarrow$ & com$\downarrow$ & acc$\downarrow$ & com$\downarrow$ & & \\
        \midrule
        w/o depth~\cite{park2024depthprompt}&3.37&3.93&4.01&4.61&3.58&3.97&7.26&8.25&5.82&6.52&6.98&7.72&6.98&6.92&4.26&6.09&-&-\\
        \cellcolor{Salmon}Midas~\cite{Ranftl2022midas}& 3.25&3.63&3.59&4.12&3.49&3.78&8.09&9.04&6.02&7.08&\underline{4.63}&\textbf{6.19}&\textbf{4.93}&\underline{5.40}&\underline{3.95}&\textbf{5.71}&\textcolor{PineGreen}{+2.32}&5\\
        \cellcolor{Salmon}DAV2-Rel~\cite{yang2024depthanythingv2} &3.30&3.92&\underline{3.52}&\textbf{3.85}&\textbf{3.28}&\textbf{3.59}&\underline{6.16}&\underline{6.94}&5.78&6.62&6.55&7.09&7.00&6.43&4.26&6.09&\textcolor{PineGreen}{+\textbf{10.00}}&\textbf{1}\\
        \cellcolor{Goldenrod}DAV2-Met~\cite{yang2024depthanythingv2}&3.22&\textbf{3.39}&\textbf{3.48}&\underline{3.98}&3.47&3.87&8.58&9.64&4.59&\underline{5.40}&6.38&7.43&6.13&5.59&3.98&6.29&\textcolor{PineGreen}{+1.95}&6\\
        \cellcolor{Goldenrod}Metric3DV2~\cite{hu2024metric3d}&\textcolor{gray}{3.48}&\textcolor{gray}{3.64}&\textcolor{gray}{3.45}&\textcolor{gray}{3.93}&\textcolor{gray}{3.73}&\textcolor{gray}{4.09}&\textcolor{gray}{9.55}&\textcolor{gray}{10.53}&\textcolor{gray}{5.82}&\textcolor{gray}{6.41}&\textcolor{gray}{5.20}&\textcolor{gray}{6.67}&\textcolor{gray}{6.73}&\textcolor{gray}{6.78}&\textcolor{gray}{4.51}&\textcolor{gray}{6.65}& \textcolor{Maroon}{-4.19} & -\\
        \cellcolor{Goldenrod}UniDepth~\cite{piccinelli2024unidepth}&\underline{3.11}&3.49&3.73&4.38&3.80&4.06&\textbf{5.96}&\textbf{6.91}&5.05&6.05&6.48&7.41&5.83&5.95&4.60&6.76&\textcolor{PineGreen}{+\underline{7.08}}&\textbf{2}\\
        \cellcolor{Apricot}Marigold~\cite{ke2024repurposing}&\textbf{3.01}&3.67&3.77&4.07&3.70&4.00&7.07&7.93&6.23&7.01&4.83&6.43&6.32&6.26&4.52&6.79&\textcolor{PineGreen}{+4.67}&4\\
        \cellcolor{Apricot}GenPercept~\cite{xu2024genpercept}&3.28&\underline{3.47}&3.77&4.34&\underline{3.33}&\underline{3.73}&7.06&7.65&\textbf{4.14}&\textbf{5.06}&\textbf{4.38}&\underline{6.35}&\underline{5.30}&\textbf{5.05}&4.40&6.20&\textcolor{PineGreen}{+6.16}&3 \\
        \cellcolor{LimeGreen}MoGe~\cite{wang2024moge}&3.26&3.67&3.67&4.23&3.89&4.33&8.86&9.83&\underline{4.55}&5.58&5.68&6.73&6.40&6.32&\textbf{3.92}&\underline{5.98} & \textcolor{Maroon}{-4.04} &7 \\
        
        Rendered&\textcolor{gray}{3.00}&\textcolor{gray}{3.29}&\textcolor{gray}{3.69}&\textcolor{gray}{4.41}&\textcolor{gray}{4.14}&\textcolor{gray}{4.47}&\textcolor{gray}{5.57}&\textcolor{gray}{6.85}&\textcolor{gray}{5.95}&\textcolor{gray}{6.75}&\textcolor{gray}{5.91}&\textcolor{gray}{7.91}&\textcolor{gray}{6.64}&\textcolor{gray}{6.65}&\textcolor{gray}{4.01}&\textcolor{gray}{6.05}&-&-\\
        \bottomrule
    \end{tabular}
    }
    
\end{table*}

\begin{figure*}
    \centering
    \includegraphics[width=0.99\linewidth]{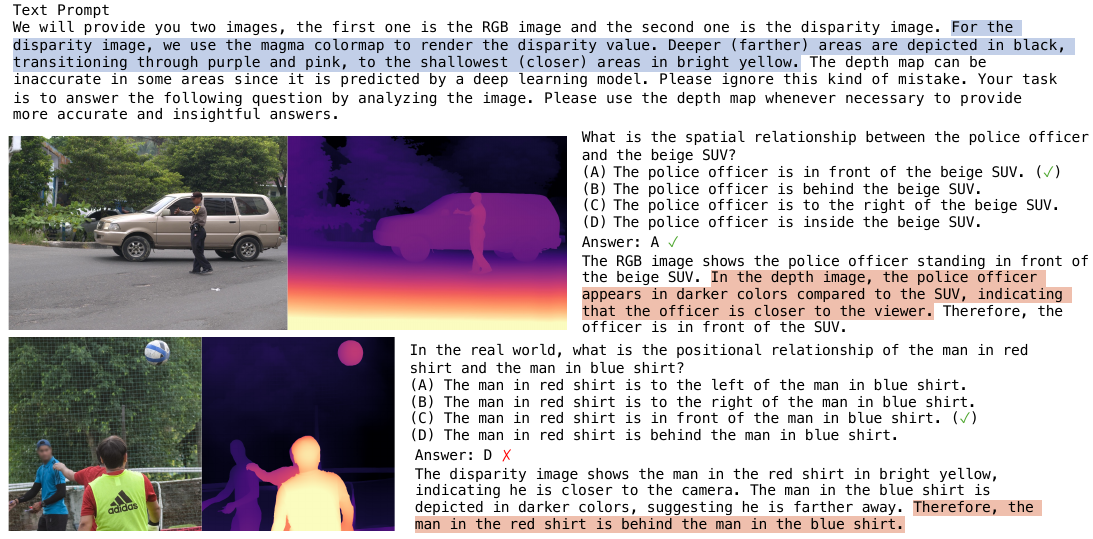}
    \caption{\textbf{Showcases of ChatGPT-4o on SpatialBot positional benchmark.} We highlight the text prompt describing rendered depth map in \colorbox[HTML]{C4CEE5}{blue} and mistakes made by ChatGPT-4o in \colorbox[HTML]{E7C1B0}{red}, respectively. In the first case, ChatGPT-4o correctly answers the question but misinterprets the depth map despite detailed prompts. As for the second one, despite correctly parsing the depth map, ChatGPT-4o provides an incorrect answer.}
    \label{fig:vlm}
\end{figure*}

\begin{table}[t]
    \centering
    \caption{\textbf{Benchmark with spatial understanding of Vision Language Model (VLM).} We evaluate the effectiveness of depth predictions from various foundation models on the SpatialBench~\cite{cai2024spatialbot}. The \textit{rank} column is omitted since all depth models perform similarly.}
    \label{tab:benchmark_vlm}
    \scalebox{0.62}{
    \begin{tabular}{L{2.3cm}|*{2}{C{1.2cm}}*{2}{C{1.2cm}}C{1.2cm}|L{2.3cm}|*{2}{C{1.2cm}}*{2}{C{1.2cm}}C{1.2cm}}
        \toprule
        Method & Pos.$\uparrow$ & Exist$\uparrow$ &Count$\uparrow$ & Reach$\uparrow$ & Size$\uparrow$ & Method & Pos.$\uparrow$ & Exist$\uparrow$ &Count$\uparrow$ & Reach$\uparrow$ & Size$\uparrow$ \\
        \midrule
        ChatGPT-4o & 64.70&95.00&80.88&54.44&31.11 & SpatialBot~\cite{cai2024spatialbot} & 61.76&75.00&92.41&51.67&28.33\\
        \cellcolor{Salmon}Midas~\cite{Ranftl2022midas} & 62.74&90.00&80.26&54.44&37.22&\cellcolor{Salmon}Midas~\cite{Ranftl2022midas} & 55.88&55.00&92.41&46.67&30.00\\
        \cellcolor{Salmon}DAV2-Rel~\cite{yang2024depthanythingv2} & 61.76&88.33&77.11&52.22&35.55&\cellcolor{Salmon}DAV2-Rel~\cite{yang2024depthanythingv2} & 55.88&60.00&93.13&46.67&30.00 \\
        \cellcolor{Goldenrod}DAV2-Met~\cite{yang2024depthanythingv2} & 61.76&86.66&80.44&59.44&38.88&\cellcolor{Goldenrod}DAV2-Met~\cite{yang2024depthanythingv2} & 55.88&65.00&93.13&45.00&28.33\\
        \cellcolor{Goldenrod}Metric3DV2~\cite{hu2024metric3d} & 62.74&88.33&79.45&59.44&28.88&\cellcolor{Goldenrod}Metric3DV2~\cite{hu2024metric3d} & 58.82&55.00&93.13&50.00&28.33\\
        \cellcolor{Goldenrod}UniDepth~\cite{piccinelli2024unidepth} & 64.70&93.33&80.55&62.22&37.77&\cellcolor{Goldenrod}UniDepth~\cite{piccinelli2024unidepth} & 58.82&60.00&92.41&53.33&28.33\\
        \cellcolor{Apricot}Marigold~\cite{ke2024repurposing} & 57.84&83.33&80.68&58.88&31.66&\cellcolor{Apricot}Marigold~\cite{ke2024repurposing} & 55.88&60.00&93.13&46.67&30.00\\
        \cellcolor{Apricot}GenPercept~\cite{xu2024genpercept} & 60.78&85.00&81.03&57.77&37.77&\cellcolor{Apricot}GenPercept~\cite{xu2024genpercept} & 55.88&65.00&93.13&48.33&28.33\\
        \cellcolor{LimeGreen}MoGe~\cite{wang2024moge} &60.78&85.00&79.06&56.11&33.33&\cellcolor{LimeGreen}MoGe~\cite{wang2024moge} &55.88&60.00&93.13&50.00&28.33\\
        \bottomrule
    \end{tabular}
    }
    
\end{table}


\noindent\textbf{Feed-Forward Monocular 3DGS.} Tab.~\ref{tab:benchmark_ffgs} shows the benchmark results. DAV2-Met achieves better performance compared with DAV2-Rel, suggesting that metric depth properties are beneficial for novel view synthesis tasks in real 3D environments. MiDaS~\cite{Ranftl2022midas}, despite being an older method, performs remarkably well with a rank of 1. DAV2-Rel also achieves strong results but slightly underperforms compared to MiDaS. Most metric depth methods, except for DAV2-Met and affine-invariant depth methods, fail to improve the baseline.

\noindent\textbf{Simultaneous Localization and Mapping.} Tab.~\ref{tab:benchmark_slam} presents the SLAM results. DAV2-Rel achieves the best results with a promising gap with other methods, indicating a superior potential for this task. UniDepth achieves the second best results, highlighting the importance of metric depth for this task. GenPercept also obtains good results, possibly due to fine-tuning on Hypersim, a similar synthetic dataset. The performance gap between GenPercept and Marigold highlights the effectiveness of its fine-tuning strategy. 


\noindent\textbf{VLM Spatial Understanding.} We use SpatialBench~\cite{cai2024spatialbot} for this task. Unlike its original purpose of benchmarking different vision-language models (VLMs), we focus on evaluating the effectiveness of different depth estimations for the same VLM. We select ChatGPT-4o and SpatialBot~\cite{cai2024spatialbot} as baseline VLMs, without and with depth inputs during training, respectively. 

Surprisingly, for both VLMs, adding depth as an additional input does not significantly improve performance, even in SpatialBot, which is trained with depth maps. All depth methods yield similar results, indicating similar effectiveness for this high-level spatial reasoning task. Fig.~\ref{fig:vlm} illustrates two cases from the positional benchmark in SpatialBench. In the first case, ChatGPT-4o correctly answers the question but misinterprets the depth map despite detailed prompts, suggesting that the training-stage with depth signals is crucial for the proper usage of depth maps. In the second case, despite correctly parsing the depth map, ChatGPT-4o provides an incorrect answer, highlighting VLMs’ current limitations in reasoning within 3D space, even when given accurate spatial information.

\section{Limitations and Future Work}
\label{sec:limitations}


While \textbf{\textit{BenchDepth}} provides a more practical evaluation framework for depth foundation models (DFMs) by leveraging downstream proxy tasks, it also introduces certain challenges. First, training on downstream tasks is computationally expensive and time-consuming. An evaluation model that can predict downstream score would alleviate this issue. Second, while we carefully selected five diverse proxy tasks to assess different aspects of depth estimation, the current set of tasks may not fully capture all potential applications of DFMs. In future work, we plan to expand BenchDepth by incorporating additional downstream tasks to further explore the capabilities.

\section{Conclusion}
\label{sec:conclusion}

We introduced \textbf{\textit{BenchDepth}}, a benchmark for evaluating depth foundation models (DFMs) through downstream proxy tasks rather than alignment-based metrics. By benchmarking \textbf{\textit{eight}} SoTA DFMs across depth completion, stereo matching, 3D scene reconstruction, SLAM, and vision-language spatial understanding, we provide a fairer and more practical assessment of their effectiveness. Our experiments reveal key insights into the performance improvement of DFMs in real-world applications as shown in Sec.~\ref{sec:intro}. By shifting depth evaluation towards real-world utility, we hope BenchDepth inspires further research, encouraging the community to rethink evaluation strategies for DFMs.



\bibliography{references}

\begin{thebibliography}{10}

\bibitem{zhang2023controlnet}
L.~Zhang, A.~Rao, and M.~Agrawala, ``Adding conditional control to text-to-image diffusion models,'' in {\em ICCV}, pp.~3836--3847, 2023.

\bibitem{li2023bevdepth}
Y.~Li, Z.~Ge, G.~Yu, J.~Yang, Z.~Wang, Y.~Shi, J.~Sun, and Z.~Li, ``Bevdepth: Acquisition of reliable depth for multi-view 3d object detection,'' in {\em AAAI}, vol.~37, pp.~1477--1485, 2023.

\bibitem{zhu2024nicerslam}
Z.~Zhu, S.~Peng, V.~Larsson, Z.~Cui, M.~R. Oswald, A.~Geiger, and M.~Pollefeys, ``Nicer-slam: Neural implicit scene encoding for rgb slam,'' in {\em 2024 International Conference on 3D Vision (3DV)}, pp.~42--52, IEEE, 2024.

\bibitem{szymanowicz2024flash3d}
S.~Szymanowicz, E.~Insafutdinov, C.~Zheng, D.~Campbell, J.~F. Henriques, C.~Rupprecht, and A.~Vedaldi, ``Flash3d: Feed-forward generalisable 3d scene reconstruction from a single image,'' {\em arXiv preprint arXiv:2406.04343}, 2024.

\bibitem{eigen2014mde}
D.~Eigen, C.~Puhrsch, and R.~Fergus, ``Depth map prediction from a single image using a multi-scale deep network,'' {\em NeurIPS}, vol.~27, 2014.

\bibitem{bhat2023zoedepth}
S.~F. Bhat, R.~Birkl, D.~Wofk, P.~Wonka, and M.~M{\"u}ller, ``Zoedepth: Zero-shot transfer by combining relative and metric depth,'' {\em arXiv preprint arXiv:2302.12288}, 2023.

\bibitem{ke2024repurposing}
B.~Ke, A.~Obukhov, S.~Huang, N.~Metzger, R.~C. Daudt, and K.~Schindler, ``Repurposing diffusion-based image generators for monocular depth estimation,'' in {\em CVPR}, pp.~9492--9502, 2024.

\bibitem{yang2024depthanythingv2}
L.~Yang, B.~Kang, Z.~Huang, Z.~Zhao, X.~Xu, J.~Feng, and H.~Zhao, ``Depth anything v2,'' {\em arXiv preprint arXiv:2406.09414}, 2024.

\bibitem{Ranftl2022midas}
R.~Ranftl, K.~Lasinger, D.~Hafner, K.~Schindler, and V.~Koltun, ``Towards robust monocular depth estimation: Mixing datasets for zero-shot cross-dataset transfer,'' {\em IEEE TPAMI}, vol.~44, no.~3, 2022.

\bibitem{wang2024moge}
R.~Wang, S.~Xu, C.~Dai, J.~Xiang, Y.~Deng, X.~Tong, and J.~Yang, ``Moge: Unlocking accurate monocular geometry estimation for open-domain images with optimal training supervision,'' {\em arXiv preprint arXiv:2410.19115}, 2024.

\bibitem{wang2025vggt}
J.~Wang, M.~Chen, N.~Karaev, A.~Vedaldi, C.~Rupprecht, and D.~Novotny, ``Vggt: Visual geometry grounded transformer,'' in {\em CVPR}, pp.~5294--5306, 2025.

\bibitem{ge2024geobench}
Y.~Ge, G.~Xu, Z.~Zhao, L.~Sun, Z.~Huang, Y.~Sun, H.~Chen, and C.~Shen, ``Geobench: Benchmarking and analyzing monocular geometry estimation models,'' {\em arXiv preprint arXiv:2406.12671}, 2024.

\bibitem{piccinelli2024unidepth}
L.~Piccinelli, Y.-H. Yang, C.~Sakaridis, M.~Segu, S.~Li, L.~Van~Gool, and F.~Yu, ``Unidepth: Universal monocular metric depth estimation,'' in {\em CVPR}, pp.~10106--10116, 2024.

\bibitem{hu2024metric3d}
M.~Hu, W.~Yin, C.~Zhang, Z.~Cai, X.~Long, H.~Chen, K.~Wang, G.~Yu, C.~Shen, and S.~Shen, ``Metric3d v2: A versatile monocular geometric foundation model for zero-shot metric depth and surface normal estimation,'' {\em IEEE TPAMI}, 2024.

\bibitem{xu2024genpercept}
G.~Xu, Y.~Ge, M.~Liu, C.~Fan, K.~Xie, Z.~Zhao, H.~Chen, and C.~Shen, ``What matters when repurposing diffusion models for general dense perception tasks?,'' {\em arXiv preprint arXiv:2403.06090}, 2024.

\bibitem{park2024depthprompt}
J.-H. Park, C.~Jeong, J.~Lee, and H.-G. Jeon, ``Depth prompting for sensor-agnostic depth estimation,'' in {\em CVPR}, pp.~9859--9869, 2024.

\bibitem{jiang2025defom}
H.~Jiang, Z.~Lou, L.~Ding, R.~Xu, M.~Tan, W.~Jiang, and R.~Huang, ``Defom-stereo: Depth foundation model based stereo matching,'' {\em arXiv preprint arXiv:2501.09466}, 2025.

\bibitem{cheng2025monster}
J.~Cheng, L.~Liu, G.~Xu, X.~Wang, Z.~Zhang, Y.~Deng, J.~Zang, Y.~Chen, Z.~Cai, and X.~Yang, ``Monster: Marry monodepth to stereo unleashes power,'' {\em arXiv preprint arXiv:2501.08643}, 2025.

\bibitem{achiam2023gpt}
J.~Achiam, S.~Adler, S.~Agarwal, L.~Ahmad, I.~Akkaya, F.~L. Aleman, D.~Almeida, J.~Altenschmidt, S.~Altman, S.~Anadkat, {\em et~al.}, ``Gpt-4 technical report,'' {\em arXiv preprint arXiv:2303.08774}, 2023.

\bibitem{li2023blip}
J.~Li, D.~Li, S.~Savarese, and S.~Hoi, ``Blip-2: Bootstrapping language-image pre-training with frozen image encoders and large language models,'' in {\em International conference on machine learning}, pp.~19730--19742, PMLR, 2023.

\bibitem{he2020moco}
K.~He, H.~Fan, Y.~Wu, S.~Xie, and R.~Girshick, ``Momentum contrast for unsupervised visual representation learning,'' in {\em CVPR}, pp.~9729--9738, 2020.

\bibitem{oquab2023dinov2}
M.~Oquab, T.~Darcet, T.~Moutakanni, H.~Vo, M.~Szafraniec, V.~Khalidov, P.~Fernandez, D.~Haziza, F.~Massa, A.~El-Nouby, {\em et~al.}, ``Dinov2: Learning robust visual features without supervision,'' {\em arXiv preprint arXiv:2304.07193}, 2023.

\bibitem{xu2023igev}
G.~Xu, X.~Wang, X.~Ding, and X.~Yang, ``Iterative geometry encoding volume for stereo matching,'' in {\em CVPR}, pp.~21919--21928, 2023.

\bibitem{zuo2024towards}
Y.~Zuo, K.~Kayan, M.~Wang, K.~Jeon, J.~Deng, and T.~L. Griffiths, ``Towards foundation models for 3d vision: How close are we?,'' {\em arXiv preprint arXiv:2410.10799}, 2024.

\bibitem{roberts2021hypersim}
M.~Roberts, J.~Ramapuram, A.~Ranjan, A.~Kumar, M.~A. Bautista, N.~Paczan, R.~Webb, and J.~M. Susskind, ``Hypersim: A photorealistic synthetic dataset for holistic indoor scene understanding,'' in {\em ICCV}, pp.~10912--10922, 2021.

\bibitem{li2024patchrefiner}
Z.~Li, S.~F. Bhat, and P.~Wonka, ``Patchrefiner: Leveraging synthetic data for real-domain high-resolution monocular metric depth estimation,'' {\em arXiv preprint arXiv:2406.06679}, 2024.

\bibitem{rombach2022sd}
R.~Rombach, A.~Blattmann, D.~Lorenz, P.~Esser, and B.~Ommer, ``High-resolution image synthesis with latent diffusion models,'' in {\em CVPR}, pp.~10684--10695, 2022.

\bibitem{geiger2013vkitti}
A.~Geiger, P.~Lenz, C.~Stiller, and R.~Urtasun, ``Vision meets robotics: The kitti dataset,'' {\em The international journal of robotics research}, vol.~32, no.~11, pp.~1231--1237, 2013.

\bibitem{silberman2012nyu}
N.~Silberman, D.~Hoiem, P.~Kohli, and R.~Fergus, ``Indoor segmentation and support inference from rgbd images,'' in {\em ECCV}, pp.~746--760, Springer, 2012.

\bibitem{geiger2012kitti}
A.~Geiger, P.~Lenz, and R.~Urtasun, ``Are we ready for autonomous driving? the kitti vision benchmark suite,'' in {\em CVPR}, pp.~3354--3361, IEEE, 2012.

\bibitem{cordts2016cityscapes}
M.~Cordts, M.~Omran, S.~Ramos, T.~Rehfeld, M.~Enzweiler, R.~Benenson, U.~Franke, S.~Roth, and B.~Schiele, ``The cityscapes dataset for semantic urban scene understanding,'' in {\em CVPR}, pp.~3213--3223, 2016.

\bibitem{li2023depthformer}
Z.~Li, Z.~Chen, X.~Liu, and J.~Jiang, ``Depthformer: Exploiting long-range correlation and local information for accurate monocular depth estimation,'' {\em Machine Intelligence Research}, pp.~1--18, 2023.

\bibitem{bhat2021adabins}
S.~F. Bhat, I.~Alhashim, and P.~Wonka, ``Adabins: Depth estimation using adaptive bins,'' in {\em CVPR}, pp.~4009--4018, 2021.

\bibitem{li2023patchfusion}
Z.~Li, S.~F. Bhat, and P.~Wonka, ``Patchfusion: An end-to-end tile-based framework for high-resolution monocular metric depth estimation,'' {\em arXiv preprint arXiv:2312.02284}, 2023.

\bibitem{chen2016diw}
W.~Chen, Z.~Fu, D.~Yang, and J.~Deng, ``Single-image depth perception in the wild,'' {\em NeurIPS}, vol.~29, 2016.

\bibitem{fu2018dorn}
H.~Fu, M.~Gong, C.~Wang, K.~Batmanghelich, and D.~Tao, ``Deep ordinal regression network for monocular depth estimation,'' in {\em CVPR}, pp.~2002--2011, 2018.

\bibitem{li2022binsformer}
Z.~Li, X.~Wang, X.~Liu, and J.~Jiang, ``Binsformer: Revisiting adaptive bins for monocular depth estimation,'' {\em arXiv preprint arXiv:2204.00987}, 2022.

\bibitem{kaplan2020scaling}
J.~Kaplan, S.~McCandlish, T.~Henighan, T.~B. Brown, B.~Chess, R.~Child, S.~Gray, A.~Radford, J.~Wu, and D.~Amodei, ``Scaling laws for neural language models,'' {\em arXiv preprint arXiv:2001.08361}, 2020.

\bibitem{yang2024depthanything}
L.~Yang, B.~Kang, Z.~Huang, X.~Xu, J.~Feng, and H.~Zhao, ``Depth anything: Unleashing the power of large-scale unlabeled data,'' {\em arXiv preprint arXiv:2401.10891}, 2024.

\bibitem{wang2024dust3r}
S.~Wang, V.~Leroy, Y.~Cabon, B.~Chidlovskii, and J.~Revaud, ``Dust3r: Geometric 3d vision made easy,'' in {\em CVPR}, pp.~20697--20709, 2024.

\bibitem{cong2025e3dbench}
W.~Cong, Y.~Liang, Y.~Zhang, Z.~Yang, Y.~Wang, B.~Ivanovic, M.~Pavone, C.~Chen, Z.~Wang, and Z.~Fan, ``E3d-bench: A benchmark for end-to-end 3d geometric foundation models,'' {\em arXiv preprint arXiv:2506.01933}, 2025.

\bibitem{lawson1995solving}
C.~L. Lawson and R.~J. Hanson, {\em Solving least squares problems}.
\newblock SIAM, 1995.

\bibitem{heath2018scientific}
M.~T. Heath, {\em Scientific computing: an introductory survey, revised second edition}.
\newblock SIAM, 2018.

\bibitem{kerbl20233dgs}
B.~Kerbl, G.~Kopanas, T.~Leimk{\"u}hler, and G.~Drettakis, ``3d gaussian splatting for real-time radiance field rendering.,'' {\em ACM Trans. Graph.}, vol.~42, no.~4, pp.~139--1, 2023.

\bibitem{xie2022neural}
Y.~Xie, T.~Takikawa, S.~Saito, O.~Litany, S.~Yan, N.~Khan, F.~Tombari, J.~Tompkin, V.~Sitzmann, and S.~Sridhar, ``Neural fields in visual computing and beyond,'' in {\em Computer Graphics Forum}, vol.~41, pp.~641--676, Wiley Online Library, 2022.

\bibitem{cai2024spatialbot}
W.~Cai, I.~Ponomarenko, J.~Yuan, X.~Li, W.~Yang, H.~Dong, and B.~Zhao, ``Spatialbot: Precise spatial understanding with vision language models,'' {\em arXiv preprint arXiv:2406.13642}, 2024.

\bibitem{martingarcia2024diffusione2eft}
G.~M. Garcia, K.~A. Zeid, C.~Schmidt, D.~de~Geus, A.~Hermans, and B.~Leibe, ``Fine-tuning image-conditional diffusion models is easier than you think,'' {\em arXiv preprint arXiv:2409.11355}, 2024.

\bibitem{mayer2016sceneflow}
N.~Mayer, E.~Ilg, P.~Hausser, P.~Fischer, D.~Cremers, A.~Dosovitskiy, and T.~Brox, ``A large dataset to train convolutional networks for disparity, optical flow, and scene flow estimation,'' in {\em CVPR}, pp.~4040--4048, 2016.

\bibitem{scharstein2014middlebury}
D.~Scharstein, H.~Hirschm{\"u}ller, Y.~Kitajima, G.~Krathwohl, N.~Ne{\v{s}}i{\'c}, X.~Wang, and P.~Westling, ``High-resolution stereo datasets with subpixel-accurate ground truth,'' in {\em Pattern Recognition: 36th German Conference, GCPR 2014, M{\"u}nster, Germany, September 2-5, 2014, Proceedings 36}, pp.~31--42, Springer, 2014.

\bibitem{schops2017eth3d}
T.~Schops, J.~L. Schonberger, S.~Galliani, T.~Sattler, K.~Schindler, M.~Pollefeys, and A.~Geiger, ``A multi-view stereo benchmark with high-resolution images and multi-camera videos,'' in {\em CVPR}, pp.~3260--3269, 2017.

\bibitem{ronneberger2015unet}
O.~Ronneberger, P.~Fischer, and T.~Brox, ``U-net: Convolutional networks for biomedical image segmentation,'' 2015.

\bibitem{zhou2018stereo}
T.~Zhou, R.~Tucker, J.~Flynn, G.~Fyffe, and N.~Snavely, ``Stereo magnification: Learning view synthesis using multiplane images,'' {\em arXiv preprint arXiv:1805.09817}, 2018.

\bibitem{straub2019replica}
J.~Straub, T.~Whelan, L.~Ma, Y.~Chen, E.~Wijmans, S.~Green, J.~J. Engel, R.~Mur-Artal, C.~Ren, S.~Verma, {\em et~al.}, ``The replica dataset: A digital replica of indoor spaces,'' {\em arXiv preprint arXiv:1906.05797}, 2019.

\end{thebibliography}
\bibliographystyle{ieeetr}

\end{document}